\documentclass[journal]{journal}
\usepackage{amsmath,amssymb,amsfonts}
\usepackage[ruled,vlined]{algorithm2e}
\usepackage{booktabs}
\usepackage{caption}
\usepackage{capt-of}
\usepackage{comment}
\usepackage{color}
\usepackage{float}
\usepackage{graphicx}
\usepackage[pdftex]{hyperref}
\usepackage{multirow}
\usepackage{nameref}
\usepackage{subfigure}
\usepackage{tabularx}
\usepackage{textcomp}
\usepackage[noadjust]{cite}
\def\BibTeX{{\rm B\kern-.05em{\sc i\kern-.025em b}\kern-.08em
		T\kern-.1667em\lower.7ex\hbox{E}\kern-.125emX}}

\ifdefined\METALABELNET
\newcommand\acctest{93.2}
\newcommand\acctrain{70.1}
\newcommand\acctrainmeta{81.2}
\else
\newcommand\acctest{91.4}
\newcommand\acctrain{86.23}
\newcommand\acctrainmeta{72.7}
\fi

\ifCLASSINFOpdf
\else
\fi
\begin{document}
%
\title{Learning from Small Amount of Medical Data with Noisy Labels: A Meta-Learning Approach}
%
%
%

\author{G\"{o}rkem Algan,
Ilkay Ulusoy,
\c{S}aban G\"{o}n\"{u}l,
Banu Turgut,
Berker Bakbak,
\thanks{G. Algan is PhD. student in the Department
of Electrical and Electronics Engineering, Middle East Technical University and works at ASELSAN. He is the corresponding author of this article.
e-mail: gorkemalgan@gmail.com.}
\thanks{I. Ulusoy is with the Department of Electrical and Electronics Engineering, Middle East Technical University, Ankara}
\thanks{\c{S}. G\"{o}n\"{u}l is with the Department of Ophthalmology, Faculty of Medicine, Selcuk University, Konya}
\thanks{B. Turgut is with the Department of Ophthalmology, Faculty of Medicine, Selcuk University, Konya}
\thanks{B. Bakbak is with the Department of Ophthalmology, Faculty of Medicine, Selcuk University, Konya}
}

%
%

\markboth{Journal of \LaTeX\ Class Files,~Vol.~6, No.~1, January~2007}%
{Shell \MakeLowercase{\textit{et al.}}: Bare Demo of IEEEtran.cls for Journals}
%



\maketitle
\thispagestyle{empty}

\begin{abstract}
  Computer vision systems recently made a big leap thanks to deep neural networks. However, these systems require correctly labeled large datasets in order to be trained properly, which is very difficult to obtain for medical applications. Two main reasons for label noise in medical applications are the high complexity of the data and conflicting opinions of experts. Moreover, medical imaging datasets are commonly tiny, which makes each data very important in learning. As a result, if not handled properly, label noise significantly degrades the performance. Therefore, a label-noise-robust learning algorithm that makes use of the \textit{meta-learning} paradigm is proposed in this article. The proposed solution is tested on retinopathy of prematurity (ROP) dataset with a very high label noise of 68\%. Results show that the proposed algorithm significantly improves the classification algorithm's performance in the presence of noisy labels.  
\end{abstract}


%
\IEEEpeerreviewmaketitle

\section{Introduction} \label{introduction}
Compared to its alternatives, deep networks are considered to have an impressive ability to generalize.
Nonetheless, these powerful models are still prone to memorize even complete random noise \cite{zhang2016understanding,krueger2017deep,arpit2017closer}.
Preventing undesired memorization of data becomes an even more important step in the presence of label noise.

Medical data is much more complicated than other standard real-world datasets and requires a high level of expertise. However, even for the experts, it is hard to correctly label data. For example, on the ROP dataset annotated by three experts, there exists 27\% of conflict \cite{wallace2008agreement}. Another work on the ROP dataset validates the same observation with eight experts \cite{campbell2017plus}. To cope with label noise, one can simply remove suspicious samples and train the network with clean data. However, medical datasets are commonly small due to the expensive cost of collecting and annotating data. Also, there are privacy concerns too. Therefore, it is crucial to make the best use of each data during training networks in the medical data domain. As a result, label-noise-robust learning is a very critical milestone for obtaining trustworthy automated diagnosis systems.

\begin{figure}[t]
    \centering
    \includegraphics[width=\columnwidth]{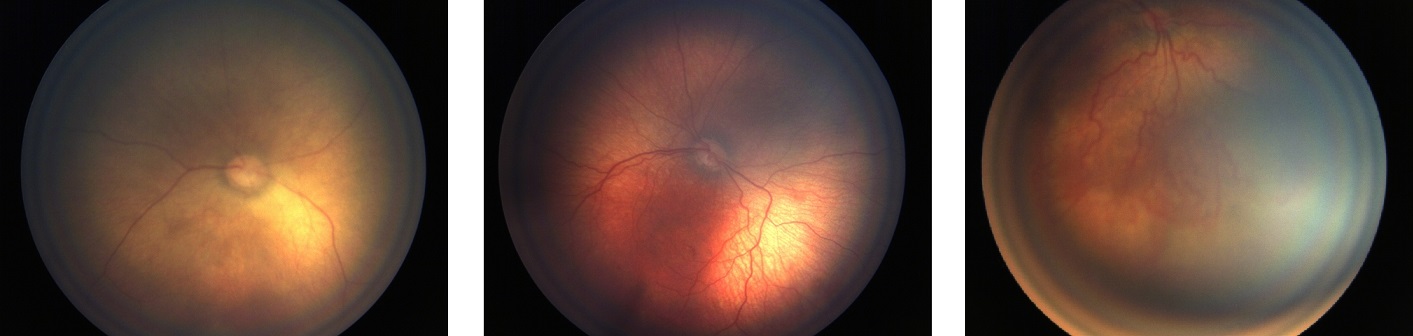}
    \caption{Retina images for different stages of ROP disease. From left to right stages are as follows; normal, pre-plus and plus.}
    \label{fig:rop_stages}
\end{figure} 

There are various approaches against label noise in the literature \cite{algan2019image}. Some works model the noise as a noisy channel between model predictions and given labels \cite{sukhbaatar2014training,sukhbaatar2014learning,bekker2016training,patrini2017making}. 
However, this is a probabilistic approach and requires a large amount of data to converge to the optimal solution, which is not the case in medical datasets. Some other works try to design noise-robust loss functions \cite{ghosh2017robust, wang2019symmetric, zhang2018generalized,wang2019improved},but it is shown that they are still affected by noise. Another trend is to increase the effect of clean samples on learning by either picking only confidently clean samples \cite{jiang2017mentornet,han2018progressive,reed2014training,han2018co} or employing a weighting scheme that would up-weight these noise-free samples \cite{wang2018iterative}. These methods are inclined to continuously learn from easy samples since they are more confidently regarded as noise-free. As a result, their learning rate is low, and they miss the valuable information from hard informative samples. One simplified approach is to cleanse noisy labels iteratively during training \cite{yi2019probabilistic}. Nevertheless, these methods are prone to re-label outliers, which are very hard to differentiate from correctly labeled hard informative samples.

All of the methods above are tested on either toy datasets (MNIST, CIFAR10 etc.) with random synthetic noise or massive datasets collected from the web. However, the medical domain is more difficult since the noise is not merely random synthetic noise, and the amount of data is much smaller than datasets collected from the web. Some pioneer works in the field of learning from medical data with noisy labels are \cite{dgani2018training,xue2019robust}, in which they used conventional learning approaches. However, in recent years, \textit{meta-learning} approaches are shown to be very effective, especially in small datasets.

\section{Meta-Learning Approach}
A conventional machine learning approach consists of four components: input data, model, loss function, and hyper-parameters. Iteratively, the model makes predictions upon the data. Then, its accuracy is calculated with the loss function. Afterward, by using stochastic gradient descent (SGD) algorithm, model parameters are updated. As a result, best model parameters are not calculated by hand, but instead, they are learned by the SGD algorithm. Besides learned model parameters, this whole process is tuned by hyper-parameters. These are special parameters that directly affect the performance of the learning algorithm (learning rate, total training duration, etc.), and they are not learned but adjusted by the algorithm designer.

\begin{figure}[h]
    \centering
    \includegraphics[width=\columnwidth]{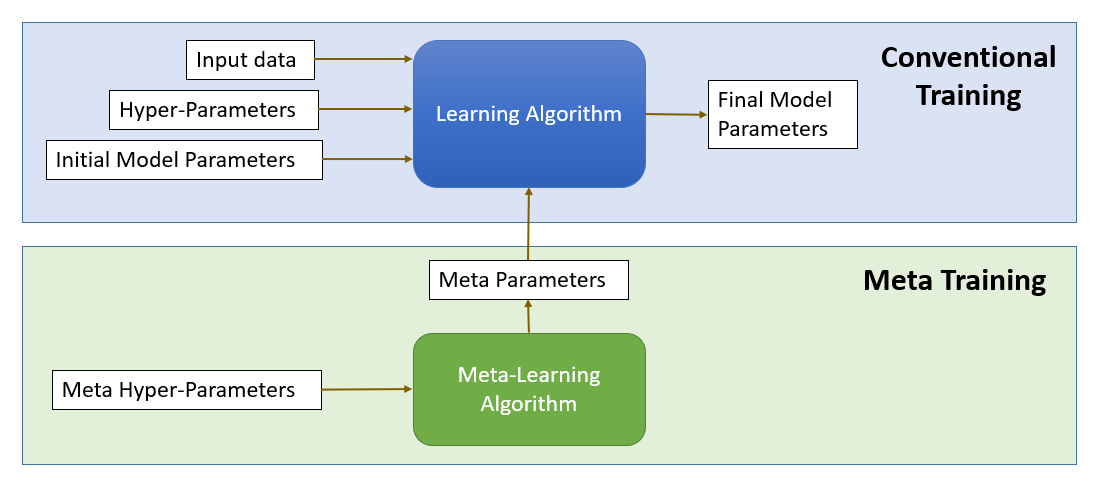}
    \caption{Illustration of meta-learning framework. Upper box represents the conventional training loop, where traditional machine learning algorithms are applied. Meta-training loop aims to optimize the conventional training loop by setting up the optimal meta-parameter set. For that purpose, it also employs a learning algorithm, using the feedback coming from the conventional training loop.}
    \label{fig:metalearning}
\end{figure}

On contrary to conventional learning, meta-learning aims to learn hyper-parameters as well. Therefore, meta-learning approaches commonly have two training loops: 1) conventional training loop that trains the base classifier with conventional machine learning techniques 2) meta training loop that optimizes meta-parameters (e.g. hyperparameters) to make conventional learning more effective. By learning hyper-parameters, meta-learning optimizes the conventional learning algorithm. Therefore, meta-learning is also called learning to learn. This architecture is illustrated in \autoref{fig:metalearning}.

In this work, the methodology of \cite{algan2020meta} is followed, which employs a meta-learning approach and has given the current state-of-the-art result on training in the presence of noisy labels. In this approach, meta-parameters are noisy labels, and the meta-learning objective is to provide noise-robust learning to the base classifier. Our algorithm is tailored to make the best use of a small verified clean data (we call this data meta-data) to extract noise-free knowledge from the noisy training data. 

The proposed framework adopts the meta-learning framework with two learning objectives as conventional training and meta training. In the conventional training stage, the base classifier is trained on training data with their predicted soft labels (instead of given noisy labels). 
\ifdefined\METALABELNET
In the meta training stage, a small multi-layer perceptron (MLP) network is trained to find optimal soft-labels to feed the conventional training phase.
\else
In the meta training stage, soft-labels are generated according to meta-objective.
\fi
It can be said that \textit{the proposed meta objective seeks optimal soft-labels for each noisy data so that base classifier trained on them would give the best performance on clean meta-data}. The proposed algorithm differs from label noise cleansing methods in a way that it is not searching for clean hard-labels but rather searching for optimal soft-labels that would provide the most noise-robust training. In order to verify the performance of the proposed algorithm, it is tested on an extremely noisy ROP dataset, which is illustrated in \autoref{fig:rop_stages}.

\section{The Proposed Method} \label{method}

\begin{figure*}[h]
    \centering
\ifdefined\METALABELNET
    \includegraphics[width=\textwidth]{meta_label_net.png}
    \caption{Illustration taken from \cite{} which visualizes the meta-training part of the training process. $x_n,y_n$ represents the training data and label pair. $x_m,y_m$ represents the meta-data and label pair. Feature extractor is the pre-softmax output of the classifier trained at warm-up phase of the training.}
\else
    \includegraphics[width=\textwidth]{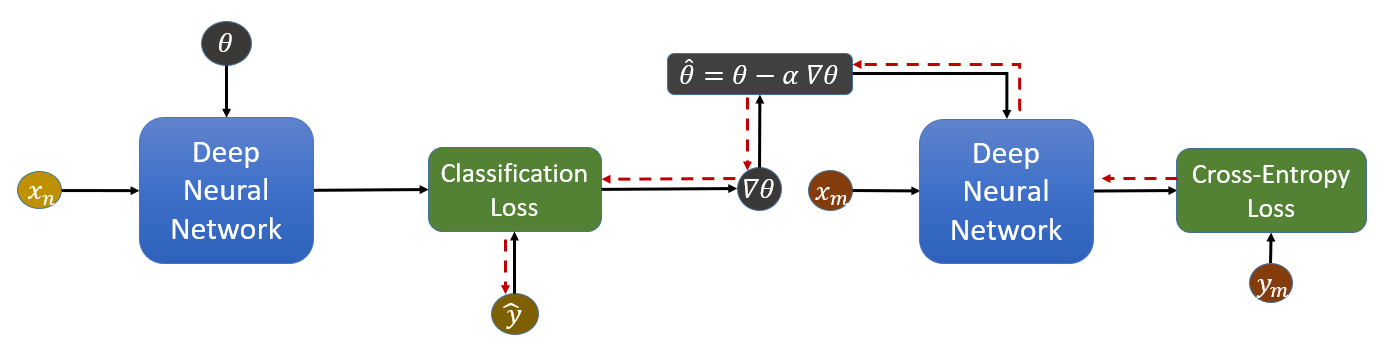}
    \caption{Meta-training part of the training process. $x_n$ represents the training data. $\hat{y}$ represents generated soft-labels. $x_m,y_m$ represents the meta-data and label pair. Notice that two deep neural networks are actually the same network, but with different parameter set. }
\fi
    \label{fig:overall}
\end{figure*}

\subsection{Problem Statement}
In supervised learning there exists a clean dataset, which consists of data instances and corresponding hard-labels. In hard-label representation, each instance belongs to only one class. In the presence of the label noise, we have the same data instances but with corrupted labels. Therefore, in the presence of the label noise, network is trained on noisy dataset while aiming to obtain the best classifier for clean dataset. However, training on noisy data directly would yield a sub-optimal model for clean dataset. As a result, one needs to consider the existence of noisy labels when training on an imperfect dataset. 

Both clean and noisy datasets are defined over hard label space $Y^h$. Differently, in this work, we are seeking optimal label distribution, which is defined over soft-label space $Y^s$. Unlike hard-labels, soft-labels have non-zero values for all classes, which lets them to define not only the corresponding class but also the similar classes. For example, in a digit classification task, "1" is more similar to "7" than "0". This relation can be expressed with soft-labels but not with hard-labels. Therefore, soft labels contain more information about the data, which is better for training. As a result, unlike aiming to find noise-free hard-labels in $Y^h$, the proposed algorithm seeks optimal soft-labels in $Y^s$.

\subsection{Learning with the Proposed Method} \label{proposedmethod}
There are two consecutive stages in training: meta training and conventional training. In the first stage, a small MLP network is trained on meta objective to find optimal soft labels. In the second stage, the base classifier is trained on these soft-labels. These two stages are repeated consecutively during the training. 

\textbf{Meta training: }Overall flowchart of meta-training stage is illustrated in \autoref{fig:overall}. 
\ifdefined\METALABELNET
First, we encode each training data instance to a feature vector by a feature extractor network. Then, the meta-network produces soft-label predictions $\hat{y}$ depending on these feature encodings. 
\else
First, soft-label predictions are generated. In the first epoch, label predictions are set as $\hat{y}=K*y$ where $K$ is a large number such as 10. In the later epochs, label-predictions are updated iteratively.
\fi
Afterward, updated model parameters are calculated $\hat{\theta}$ by using these labels with stochastic gradient descent (SGD) optimizer as follows

\begin{equation}
    \hat{\theta} = \theta^{(t)} - \alpha \nabla_\theta \mathcal{L}_{KL}(f_\theta(x),\hat{y}^{(t)})\Biggr|_{\theta^{(t)}}
    \label{eq:thetahat}
\end{equation}

We used KL-divergence loss $\mathcal{L}_{KL}$ as loss function since it is a defacto loss function for soft-labels. Then using these updated parameters, the loss on meta-data is calculated. Since meta-data consists of clean hard labels, categorical cross-entropy loss $\mathcal{L}_{cce}$ is used to calculate meta-loss. Then the meta loss is backpropagated all the way back for updating soft-labels. This is more like asking the question; \textit{what would be the initial label predictions so that the base classifier trained on it ($\hat{\theta}$) would give the minimum loss on clean meta-data}. Intuitively, the answer is the noise-free soft labels of the corresponding data. Therefore, the meta-objective is to find the least noise-affected soft-label representation of data.
\ifdefined\METALABELNET

However, it should be prevented that label predictions to go delusional and generate totally different labels from given labels. Even though there exists noisy labels, there are clean labels too and we want to make use of them. Therefore a consistency loss $\mathcal{L}_{con}$ is defined, which is the log-likelihood of generated labels and given noisy labels. This loss ensures that generated labels do not diverge too much from the given labels. Finally, meta-network parameters are updated with SGD on the meta loss and consistency loss

\begin{equation}
    \phi^{(t+1)} = \phi^{(t)} - \nabla_\phi (\beta \mathcal{L}_{cce}(f_{\hat{\theta}}(x), y)  + \gamma \mathcal{L}_{con}(m_\phi(x), \tilde{y}))\Biggr|_{\phi^{(t)}}
    \label{eq:phi}
\end{equation}
\else
Label predictions are updated with SGD on the meta loss, which is calculated over meta-data.

\begin{equation}
    \hat{y}^{(t+1)} = \hat{y}^{(t)} - \nabla_{\hat{y}} \beta \mathcal{L}_{cce}(f_{\hat{\theta}}(x), y)\Biggr|_{\hat{y}^{(t)}}
    \label{eq:phi}
\end{equation}
\fi

\textbf{Conventional training: }At this point the soft-label predictions are generated. Therefore, the base classifier is trained on these soft-labels instead of given noisy labels. Same as before, KL-divergence loss $\mathcal{L}_{KL}$ is used for loss function. Furthermore, soft-labels can peak at multiple locations, but it is desired to have network prediction to peak at only one value. Therefore, entropy loss $\mathcal{L}_e$ is defined as follows, which forces network predictions to peak at only one value. 

\begin{equation}
    \mathcal{L}_e(f_\theta(x)) = - \dfrac{1}{N} \sum_{i=1}^{N} \sum_{j=1}^{C} f^j_\theta(x_i) \log(f^j_\theta(x_i))
\end{equation}

Finally, the base-classifier is trained by using these two loss functions with SGD as follows

\begin{equation}
    \theta^{(t+1)} = \theta^{(t)} - \lambda \nabla_\theta \left(\mathcal{L}_{KL}(f_\theta(x),\hat{y}^{(t+1)}) + \mathcal{L}_{e}(f_\theta(x))\right)\Biggr|_{\theta^{(t)}}
    \label{eq:train}
\end{equation}

\ifdefined\METALABELNET
Notice that we have four different learning rates $\alpha, \beta, \gamma$ and $\lambda$ for each step.
\else
Notice that there are three different learning rates $\alpha, \beta$ and $\lambda$ for each step.
\fi

\subsection{Overall Algorithm}
It is empirically shown that, under the presence of the noise, deep networks first learn clean representations of data, and only afterward they start to overfit the noise. 
Therefore, we begin by training network on noisy data with classical cross-entropy loss as warm-up training. At the end of warm-up training, the model leverages useful information from the data. This is also beneficial for the meta-training stage. Since gradients are taken on the feedback coming from the base classifier, without any pre-training, random feedbacks coming from the base network would cause meta-objective to lead in the wrong direction. After warm-up training, the proposed algorithm is employed, as explained in Section \ref{proposedmethod}. 

\ifdefined\METALABELNET
For feature extractor, the same model state after the warm-up training without additional learning is used.
\else
\fi
\section{Experiments} \label{experiments}
\subsection{Dataset}
We collected 1947 retina images with 640x480 resolution from potential ROP patients. Each image is labeled by the same three experts to one of the following categories: normal, pre-plus and plus. From these 1947 images, only on 622 images all three experts gave the same label, which is 32\% of the dataset. We randomly picked 200 images as meta-data and 300 images as test-data from these 622 images. Remaining 1447 images are considered to be unreliable since there is no consensus on the labels by three experts. Therefore, these images are used as noisy training data. We used majority voting on training data annotations to determine training labels. For preprocessing, all images are resized to 256x256 and the center crop 224x224. For data augmentation purposes, a random horizontal flip is used.

\subsection{Pre-Training}
We used ResNet50 architecture with model parameters pre-trained on the ImageNet dataset. Our dataset is considerably small; therefore, transfer learning is applied by further pre-training network on a different but similar dataset. We used diabetic retinopathy (DR)  dataset that has 35k high-resolution retina images \cite{cuadros2009eyepacs}. A clinician has rated the existence of diabetic retinopathy on a scale of 0 to 4 as follows: no diabetic retinopathy (0), mild (1), moderate (2), severe (3), proliferative DR (4). Since main task is not classifying DR, but rather pre-train network to learn useful representation mapping, dataset is converted into binary classification task as: diabetic retinopathy (0) and no diabetic retinopathy (1). This dataset consists of high-resolution images with varying sizes. Therefore, all images are resized to 1024x1024 and then cropped 224x224 around the center of the retina image. On the resulting dataset, the model is trained for one epoch. Afterward, the final layer of the classifier is replaced with randomly initialized 2048x4 fully connected layer and softmax layer to match the ROP dataset labels.

\subsection{Training Procedure}
After pre-training on DR dataset, the proposed algorithm is employed. Stochastic gradient descent optimizer with momentum 0.9 and weight decay $10^{-4}$ is used for base classifier. We initialized learning rate as $10^{-3}$ and set it to $10^{-4}$ and $10^{-5}$ at $10^{th}$ and $20^{th}$ epochs. Total training consists of 30 epochs, in which first 10 epochs are warm-up training and rest is meta-training. During the whole training batch size of 16 is trained. 
\ifdefined\METALABELNET
It is observed that the increasing depth of the meta-network does not contribute to the overall performance. Therefore, a single layer network is used as meta-network with size NUM FEATURES (2048) x NUM CLASSES (10). Adam optimizer with $10^{-4}$ learning rate and $10^{-5}$ weight decay is used for meta network. We set the hyper-parameters as $\alpha=0.5, \beta=10^{-3}, \gamma=0.1$.
\else
We set the hyper-parameters as $K=10, \alpha=0.5, \beta=4000$.
\fi

\subsection{Evaluation of the results}
In order to show the effectiveness of the propsed algorithm, its performance is compared to classical learning with cross-entropy loss. We made three different runs, and results are provided in \autoref{results}. First, the model is directly trained on the ROP dataset without any pre-training on the DR dataset. This run achieved 86.3\% test accuracy, which is moderate but the worst performance in the leaderboard. Secondly, the network is pre-trained on the DR dataset and then move to conventional training with cross-entropy on the ROP dataset. This run resulted in 90.3\%, which is 4\% more than the previous run. Increase in the performance shows the effectiveness of pre-training on the DR dataset. Finally, the proposed learning framework is employed. Our proposed algorithm gives the best performance with \acctest\%. Another important observation is, in conventional cross-entropy loss both networks manage to get 100\% training accuracy. Considering that labels are extremely noisy, this is an undesired behavior that means the network is overfitting the data. On the other hand, in the presented algorithm, training accuracy is stuck at \acctrain\%. Therefore, it can be concluded that the proposed algorithms prevents model to overfit the noise. 
\ifdefined\METALABELNET
Moreover, when we measure the training accuracy with generated labels by meta-network, it is \acctrainmeta\%. This also indicates that generated labels have a cleaner representation of the data since it is higher than training accuracy.
\else
\fi

\begin{table}[]
    \centering
    \resizebox{\columnwidth}{!}{
    \begin{tabular}{l|l|l}
        \hline
        \multicolumn{1}{c|}{\textbf{Method}}     & \multicolumn{1}{c|}{\textbf{Train}}  & \multicolumn{1}{c}{\textbf{Test}}   \\ \hline
        Cross Entropy (no pre-train)    & 100.0\%  & 86.3\% \\
        Cross Entropy (pre-train on DR) & 100.0\%  & 90.3\% \\
        \textbf{Proposed algorithm}     & \textbf{\acctrain\%} & \textbf{\acctest\%} \\ \hline
    \end{tabular}}
    \caption{Train and test accuracies on ROP dataset.}
    \label{results}
\end{table}

\section{Discussion} \label{method}
Automated diagnosis of ROP is a trending topic, and various works are devoted to it in the literature. What differentiates the proposed method from the previous works is its capability to work with very few samples. This is due to the meta-learning approach used in the algorithm. Used meta-objective effectively utilizes a very few amounts of verified data to extract useful information from the noisy training data. It can be said that the meta-objective guides conventional learning abjective to make the best use of the limited data. The presented algorithm gives the same performance with \cite{brown2018automated,wang2019automated} even though it uses a much smaller training dataset. For example \cite{brown2018automated} uses 5.5k images and \cite{wang2019automated} uses 8.5k images. The proposed method achieved the same test performance as mentioned works by only using 1.9k images. Some other works \cite{wang2018automated,zhang2018development,huang2020automated} manage to get higher test accuracies than 94\%. However, they all use training datasets bigger than 10k images. Depending on these observations, it can be said that the presented algorithm is especially handy in the case of limited data.


\section{Conclusion} \label{conclusion}
In this work, a meta-learning based label noise robust learning algorithm is proposed. The presented method has two learning objectives as conventional training and meta training. In the meta training stage aim is to find optimum soft-label representation of the noisy training data. In the conventional training stage, the base classifier is trained with predicted soft-labels instead of given noisy labels. The presented algorithm is especially effective in the case of small data with label noise, since it can successfully leverage the noise-free information from the noisy training data by using very small clean meta-data. We tested the proposed algorithm with the ROP dataset with extreme label noise ratio of 68\%, where the proposed algorithm managed to get best test accuracy of \acctest\%.

\section{Acknowledgements}
No funding was received for conducting this study. The authors have no relevant financial or non-financial interests to disclose.

\section{Compliance with Ethical Standards}
This study was performed in line with the principles of the Declaration of Helsinki. Approval was granted by the Ethics Committee of Selcuk University, Faculty of Medicine (Date:13.09.2017, No:2017/15)

\bibliographystyle{IEEEbib}
\bibliography{references} 

\begin{thebibliography}{10}

\bibitem{zhang2016understanding}
Chiyuan Zhang, Samy Bengio, Moritz Hardt, Benjamin Recht, and Oriol Vinyals,
\newblock ``Understanding deep learning requires rethinking generalization,''
\newblock {\em arXiv preprint arXiv:1611.03530}, 2016.

\bibitem{krueger2017deep}
David Krueger, Nicolas Ballas, Stanislaw Jastrzebski, Devansh Arpit, Maxinder~S
  Kanwal, Tegan Maharaj, Emmanuel Bengio, Asja Fischer, and Aaron Courville,
\newblock ``Deep nets don't learn via memorization,''
\newblock 2017.

\bibitem{arpit2017closer}
Devansh Arpit, Stanislaw Jastrzebski, Nicolas Ballas, David Krueger, Emmanuel
  Bengio, Maxinder~S Kanwal, Tegan Maharaj, Asja Fischer, Aaron Courville,
  Yoshua Bengio, et~al.,
\newblock ``A closer look at memorization in deep networks,''
\newblock in {\em Proceedings of the 34th International Conference on Machine
  Learning-Volume 70}. JMLR. org, 2017, pp. 233--242.

\bibitem{wallace2008agreement}
David~K Wallace, Graham~E Quinn, Sharon~F Freedman, and Michael~F Chiang,
\newblock ``Agreement among pediatric ophthalmologists in diagnosing plus and
  pre-plus disease in retinopathy of prematurity,''
\newblock {\em Journal of American Association for Pediatric Ophthalmology and
  Strabismus}, vol. 12, no. 4, pp. 352--356, 2008.

\bibitem{campbell2017plus}
John~P Campbell, Jayashree Kalpathy-Cramer, Deniz Erdogmus, Susan Ostmo, Ryan
  Swan, Kemal Sonmez, RV~Paul Chan, and Michael~F Chiang,
\newblock ``Plus disease in rop: why do experts disagree, and how can we
  improve diagnosis?,''
\newblock {\em Journal of American Association for Pediatric Ophthalmology and
  Strabismus $\{$JAAPOS$\}$}, vol. 21, no. 4, pp. e5--e6, 2017.

\bibitem{algan2019image}
Görkem Algan and Ilkay Ulusoy,
\newblock ``Image classification with deep learning in the presence of noisy
  labels: A survey,''
\newblock {\em Knowledge-Based Systems}, vol. 215, pp. 106771, 2021.

\bibitem{sukhbaatar2014training}
Sainbayar Sukhbaatar, Joan Bruna, Manohar Paluri, Lubomir Bourdev, and Rob
  Fergus,
\newblock ``Training convolutional networks with noisy labels,''
\newblock {\em arXiv preprint arXiv:1406.2080}, 2014.

\bibitem{sukhbaatar2014learning}
Sainbayar Sukhbaatar and Rob Fergus,
\newblock ``Learning from noisy labels with deep neural networks,''
\newblock {\em arXiv preprint arXiv:1406.2080}, vol. 2, no. 3, pp. 4, 2014.

\bibitem{bekker2016training}
Alan~Joseph Bekker and Jacob Goldberger,
\newblock ``Training deep neural-networks based on unreliable labels,''
\newblock in {\em 2016 IEEE International Conference on Acoustics, Speech and
  Signal Processing (ICASSP)}. IEEE, 2016, pp. 2682--2686.

\bibitem{patrini2017making}
Giorgio Patrini, Alessandro Rozza, Aditya Krishna~Menon, Richard Nock, and
  Lizhen Qu,
\newblock ``Making deep neural networks robust to label noise: A loss
  correction approach,''
\newblock in {\em Proceedings of the IEEE Conference on Computer Vision and
  Pattern Recognition}, 2017, pp. 1944--1952.

\bibitem{ghosh2017robust}
Aritra Ghosh, Himanshu Kumar, and PS~Sastry,
\newblock ``Robust loss functions under label noise for deep neural networks,''
\newblock in {\em Thirty-First AAAI Conference on Artificial Intelligence},
  2017.

\bibitem{wang2019symmetric}
Yisen Wang, Xingjun Ma, Zaiyi Chen, Yuan Luo, Jinfeng Yi, and James Bailey,
\newblock ``{Symmetric Cross Entropy for Robust Learning with Noisy Labels},''
\newblock 2019.

\bibitem{zhang2018generalized}
Zhilu Zhang and Mert Sabuncu,
\newblock ``Generalized cross entropy loss for training deep neural networks
  with noisy labels,''
\newblock in {\em Advances in neural information processing systems}, 2018, pp.
  8778--8788.

\bibitem{wang2019improved}
Xinshao Wang, Elyor Kodirov, Yang Hua, and Neil~M Robertson,
\newblock ``{Improved Mean Absolute Error for Learning Meaningful Patterns from
  Abnormal Training Data},''
\newblock Tech. {R}ep., 2019.

\bibitem{jiang2017mentornet}
Lu~Jiang, Zhengyuan Zhou, Thomas Leung, Li-Jia Li, and Li~Fei-Fei,
\newblock ``Mentornet: Learning data-driven curriculum for very deep neural
  networks on corrupted labels,''
\newblock {\em arXiv preprint arXiv:1712.05055}, 2017.

\bibitem{han2018progressive}
Bo~Han, Ivor~W Tsang, Ling Chen, P~Yu Celina, and Sai-Fu Fung,
\newblock ``Progressive stochastic learning for noisy labels,''
\newblock {\em IEEE transactions on neural networks and learning systems}, ,
  no. 99, pp. 1--13, 2018.

\bibitem{reed2014training}
Scott Reed, Honglak Lee, Dragomir Anguelov, Christian Szegedy, Dumitru Erhan,
  and Andrew Rabinovich,
\newblock ``Training deep neural networks on noisy labels with bootstrapping,''
\newblock {\em arXiv preprint arXiv:1412.6596}, 2014.

\bibitem{han2018co}
Bo~Han, Quanming Yao, Xingrui Yu, Gang Niu, Miao Xu, Weihua Hu, Ivor Tsang, and
  Masashi Sugiyama,
\newblock ``Co-teaching: Robust training of deep neural networks with extremely
  noisy labels,''
\newblock in {\em Advances in Neural Information Processing Systems}, 2018, pp.
  8527--8537.

\bibitem{wang2018iterative}
Yisen Wang, Weiyang Liu, Xingjun Ma, James Bailey, Hongyuan Zha, Le~Song, and
  Shu-Tao Xia,
\newblock ``Iterative learning with open-set noisy labels,''
\newblock in {\em Proceedings of the IEEE Conference on Computer Vision and
  Pattern Recognition}, 2018, pp. 8688--8696.

\bibitem{yi2019probabilistic}
Kun Yi and Jianxin Wu,
\newblock ``Probabilistic end-to-end noise correction for learning with noisy
  labels,''
\newblock in {\em Proceedings of the IEEE Conference on Computer Vision and
  Pattern Recognition}, 2019, pp. 7017--7025.

\bibitem{dgani2018training}
Yair Dgani, Hayit Greenspan, and Jacob Goldberger,
\newblock ``Training a neural network based on unreliable human annotation of
  medical images,''
\newblock in {\em 2018 IEEE 15th International Symposium on Biomedical Imaging
  (ISBI 2018)}. IEEE, 2018, pp. 39--42.

\bibitem{xue2019robust}
Cheng Xue, Qi~Dou, Xueying Shi, Hao Chen, and Pheng-Ann Heng,
\newblock ``Robust learning at noisy labeled medical images: Applied to skin
  lesion classification,''
\newblock in {\em 2019 IEEE 16th International Symposium on Biomedical Imaging
  (ISBI 2019)}. IEEE, 2019, pp. 1280--1283.

\bibitem{algan2020meta}
G{\"o}rkem Algan and Ilkay Ulusoy,
\newblock ``Meta soft label generation for noisy labels,''
\newblock in {\em Proceedings of the 25th International Conferance on Pattern
  Recognition, ICPR, pp. 7142--7148}, 2020.

\bibitem{cuadros2009eyepacs}
Jorge Cuadros and George Bresnick,
\newblock ``Eyepacs: an adaptable telemedicine system for diabetic retinopathy
  screening,''
\newblock {\em Journal of diabetes science and technology}, vol. 3, no. 3, pp.
  509--516, 2009.

\bibitem{brown2018automated}
James~M Brown, J~Peter Campbell, Andrew Beers, Ken Chang, Susan Ostmo, RV~Paul
  Chan, Jennifer Dy, Deniz Erdogmus, Stratis Ioannidis, Jayashree
  Kalpathy-Cramer, et~al.,
\newblock ``Automated diagnosis of plus disease in retinopathy of prematurity
  using deep convolutional neural networks,''
\newblock {\em JAMA ophthalmology}, vol. 136, no. 7, pp. 803--810, 2018.

\bibitem{wang2019automated}
Yifan Wang and Yuanyuan Chen,
\newblock ``Automated recognition of retinopathy of prematurity with deep
  neural networks,''
\newblock in {\em Journal of Physics: Conference Series}. IOP Publishing, 2019,
  vol. 1187, p. 042057.

\bibitem{wang2018automated}
Jianyong Wang, Rong Ju, Yuanyuan Chen, Lei Zhang, Junjie Hu, Yu~Wu, Wentao
  Dong, Jie Zhong, and Zhang Yi,
\newblock ``Automated retinopathy of prematurity screening using deep neural
  networks,''
\newblock {\em EBioMedicine}, vol. 35, pp. 361--368, 2018.

\bibitem{zhang2018development}
Yinsheng Zhang, Li~Wang, Zhenquan Wu, Jian Zeng, Yi~Chen, Ruyin Tian, Jinfeng
  Zhao, and Guoming Zhang,
\newblock ``Development of an automated screening system for retinopathy of
  prematurity using a deep neural network for wide-angle retinal images,''
\newblock {\em IEEE access}, vol. 7, pp. 10232--10241, 2018.

\bibitem{huang2020automated}
Yo-Ping Huang, Haobijam Basanta, Eugene Yu-Chuan Kang, Kuan-Jen Chen, Yih-Shiou
  Hwang, Chi-Chun Lai, John~P Campbell, Michael~F Chiang, Robison Vernon~Paul
  Chan, Shunji Kusaka, et~al.,
\newblock ``Automated detection of early-stage rop using a deep convolutional
  neural network,''
\newblock {\em British Journal of Ophthalmology}, 2020.

\end{thebibliography}

\end{document}